\documentclass{article}
\usepackage{ijcai19}

\usepackage{times}
\usepackage{soul}
\usepackage{url}
\usepackage[hidelinks]{hyperref}
\usepackage[utf8]{inputenc}
\usepackage[small]{caption}
\usepackage{graphicx}
\usepackage{amsmath}
\usepackage{booktabs}
\usepackage{algorithm}
\usepackage{algorithmic}
\title{Hyperspectral Classification Based on 3D Asymmetric Inception Network with Data Fusion Transfer Learning}

\author{
   Haokui Zhang$^{1,2^\dag}$, Yu Liu$^{2^\dag}$, Bei Fang$^{1}$, Ying Li$^{1}$, Lingqiao Liu$^{2}$ and Ian Reid$^{2}$
    \thanks{
    \textit{
    \noindent
    \dag Haokui Zhang and Yu Liu contributed equally to this work. \newline
    $^{1}$ H. Zhang, B. Fang and Y. Li are with School of Computer Science and Engineering, Northwestern Polytechnical University, Xi'an, 710072, China
{\tt\small hkzhang1991@mail.nwpu.edu.cn}
    \newline
    $^{2}$Y. Liu L. Liu and I. Reid 
are with School of Computer Science, The University of Adelaide,
5005, North Terrace, SA 
{\tt\small yu.liu04@adelaide.edu.au}
    }
    }
}

\begin{document}

\maketitle

\begin{abstract}
Hyperspectral image(HSI) classification has been improved with convolutional neural network(CNN) in very recent years. Being different from the RGB datasets, different HSI datasets are generally captured by various remote sensors and have different spectral configurations. Moreover, each HSI dataset only contains very limited training samples and thus it is prone to overfitting when using deep CNNs. In this paper, we first deliver a 3D asymmetric inception network, AINet, to overcome the overfitting problem. With the emphasis on spectral signatures over spatial contexts of HSI data, AINet can convey and classify the features effectively. In addition, the proposed data fusion transfer learning strategy is beneficial in boosting the classification performance. Extensive experiments show that the proposed approach beat all of the state-of-art methods on several HSI benchmarks, including Pavia University, Indian Pines and Kennedy Space Center(KSC). Code can be found at: https://github.com/UniLauX/AINet
\footnote{Code is avaliable at: https://github.com/UniLauX/AINet}.

\end{abstract}

\section{Introduction}


For a long time, how to extract the useful information from HSI dataset itself is a very challenging task. At first, researchers mainly focus on extracting spectral signatures while totally ignoring the spatial contexts. Later on, there are two categories of methods to extract the spectral-spatial information from HSI dataset. The first category of methods are to extract the spectral signatures and the spatial contexts separately \cite{jia2015spectral,benediktsson2005classification}. The second category of methods are to fuse the spectral signatures and the spatial contexts first, and then extract the concatenated information later \cite{qian2013hyperspectral}. Between those two categories of methods, it proves that the second category of methods are better in improving the performance of HSI classification. 

Among traditional methods, handcrafted features usually are used, and they are expected to be discriminative and representative for capturing the characteristics of the HSI datasets. While in most cases, the extracted features are domain knowledge oriented, thus may lose some valuable details. For feature classification, support vector machine(SVM) \cite{melgani2004classification} is employed because SVM is robust at representing the high-dimensional vectors, but the capacity for representation is still limited to finite dimensions.




Since 2012, with the renaissance of deep learning, the performance of many vision tasks have been dramatically improved. The reason for the improvements can be mainly summarized to two aspects, that are deep neural network structures and huge training datasets. One of the most absorbing and significant advantage of deep learning is the ability to use CNN to extract features in different granularities. The potential of CNN has also been utilized in HSI classification, and it has been proved works better than the traditional methods. There are several different network structures have been proposed to combine CNN into HSI classification, including residual network in 2D and 3D \cite{zhong2018spectral}, deep belief network(DBN) \cite{chen2015spectral}, stacked autoencoder(SAE) \cite{chen2014deep} \textit{el al}. All of the networks beat the traditional methods a bit, and which obviously point out that the irresistible trend of deep learning is also the right direction that HSI classification should be toward to. 


Generally speaking, there are two constraints which limit the usage of state-of-the-art deep CNNs, which are already employed in vision tasks, directly into HSI classification. The first one is the different data formats between RGB image and HSI. In particular, comparing to the RGB image, which can be well represented by a 2D CNN to extract the features, it is much sensible to utilize a 3D CNN to preserve the abundant information being extracted from the spectral signatures and the spatial contexts of HSI. Moreover, all of those existing networks in HSI classification are quite shallow, almost less than five layers regarding the depth of the network layers, which is eligible comparing with the one thousand layers' ResNet \cite{he2016deep} be used in other vision tasks. The reason behind that is the very limited HSI datasets. Specifically, both the capture and the annotation are cost-expensive and labor-exhaustive. Therefore, when deep CNNs are used in HSI classification, over-fitting is likely to occur.


In this paper, we propose a AINet and use two strategies to improve the HSI classification. Our contributions can be summarized as follows: 
\begin{enumerate}
\item
Firstly, a novel deep light-weight 3D CNN
with asymmetric structure is delivered to handle HSI classification, which make it possible to use the existing small volume of HSI datasets to train the very deep neural network and fully exert the potential of CNN.  
\item
Secondly, data fusion transfer learning is exploited to conduct a better model initialization. Which is compensated for the data limitation and dramatically boost the training efficiency and classification performance.

\end{enumerate}


\section{Related Works}
\label{sec:relatedwork}
\subsection{HSI classification}

Classifying each pixel into its corresponding category is a vital problem for hyperspectral image analysis, and the applications of HSI classification covers object recognition \cite{zhang2016deep}, mineral exploitation \cite{yokoya2016potential} and other relevant research fields.

\subsubsection{Conventional HSI Classification}

Generally speaking, conventional pixel-wise HSI classification mainly focus on two aspects, that are feature extraction and feature classification. Since spectral signatures usually composed of at least hundreds of bands and carry rich information about HSI, it is important to acquire the features that are discriminative and representative for HSI dataset. Regarding to feature extraction, there are two kinds of approaches for the purpose. The first kind of approaches use linear algorithms to select the representative spectral bands, which including mainford ranking \cite{wang2016salient}, multitask joint sparse representation \cite{yuan2016hyperspectral} \textit{et al}. To sum up, features are usually the extracted spectral bands from the raw HSI dataset and the physical meaning can be retained. The second kind of approaches employ non-linear algorithms to extract the discriminative features, which including principle component analysis (PCA), identity component analysis (ICA) {et al}. The step after feature extraction is feature classification, support vector machine (SVM) is usually utilized because it is robust to the high-dimensional vectors \cite{zhang2012combining}. However, the limited capacity of SVM to simulate the distribution of spectral bands become a bottleneck for HSI classification.

Later on, with the development of technologies in remote sensing (RS), some works propose to integrate the spatial context into HSI classification. Considering the order for extracting spatial contexts, there are two branches: post-processing and spatial extraction.
As its name, the post-processing approaches first extracting the spectral and spatial information sequentially, and then using the nearby spectral pixels as the smooth priors for the target spatial pixel, and a graphical model is used to conduct the classification \cite{tarabalka2010svm}. On the contrary, the spatial extraction methods emphasize to extract a 3D cube from both the spectral and spatial dimension, and SVM is used to execute the classification. Comparing with the post-processing approaches, spatial extraction approaches can achieve better performance, which may due to the reason that post-processing approaches often lose important information during the feature extraction. In addition, both methods share two drawbacks, one is the engineering features just preserve part of the information and can not fully represent the HSI dataset, the other is the traditional methods have limited capacity to fit the abundant information hied in HSI dataset. 

\subsubsection{DL Based HSI Classification}
Usually, a CNN model is composed of at least three convolutional layers for extracting features both in low-level and high level. In specific, with bottom-layers extracting the textures and edge details, and with the top-layers extracting the abstract shape information. However, traditional methods can only extract limited low-level features. The second advantage of CNN based methods compared with traditional methods for HSI classification is that, instead of separating the feature extraction and feature classification as two steps, the CNN structure integrating the feature extraction and feature classification into one framework
through back-propagation, and since the extracted features directly contribute the final classification performance, deep learning methods achieve better performance than the traditional kernel based methods.

From the network structure perspective, there are several representative works for HSI classification. Stacked autoencoder(SAE) stack the extracted spectral and spatial features using layer-wise pretraining models \cite{chen2014deep}. Deep belief work(DBN) is also explored for HSI classification \cite{chen2015spectral}. However, both of the methods require to compress the spatial contexts as the 1D vector, which inevitably lose the important information about HSI. 2D-CNN network is directly transfered from the one used for vision tasks to HSI classification, and some researchers \cite{li2017hyperspectral,tarabalka2009spectral} propose to use two parallel 2D-CNNs to extract the spectral signatures and the spatial contexts, both of the feature extraction for the spatial contexts and the separation between spectral and spatial channels will cause information lose. Moreover, 3D-CNN \cite{chen2016deep,li2017spectral} is proposed to extract the spectral signatures and spatial contexts simultaneously from the HSI dataset, and conducting the classification on the 3D cube, but the 3D CNN models will suffer from over-fitting when the layers of the network become deeper, which mainly due to the very limited training dataset of HSI. 


\subsection{3D CNN Architectures}
The first 3D CNN network for HSI classification is proposed by Chen \textit{et al} in 2016 \cite{chen2016deep}, the ${L}_{2}$-norm regularization and dropout are used. However, the layers of the network is very shallow and it still come across the over-fitting problem when the available annotated datasets are scarce. A similar work is the spectral-spatial residual network(SSRN) \cite{zhong2018spectral}, which bring the widely used residual blocks from other vision tasks into HSI classification, and combining the batch normalization to achieve a better performance. The disadvantage of SSRN is that it do not explicitly considering the contribution difference between spectral signatures and spatial contexts. our proposed asymmetric residual network beat SSRN with huge gains, which not only benefit from the much deeper and light-weight network design, but also due to the AI unit we come up with that is tailored for HSI dataset. 
\subsection{Transfer Learning}
Comparing with the thousands of million annotated datasets used in some vision tasks, the existing available HSI annotated datasets are quite scarce. Moreover, the serve unbalance among HSI datasets of intra-class and captured from different sensors also make it challenging to train the neural network for HSI classification. In vision community, one common solution for this problem is the transfer learning. The work principle behind transfer learning is that, in deep neural network, the bottom-level and middle-level features taking up the majority of the parameters stored in the CNN model, and usually capturing the textures and edges of the objects. And those low-level features designed for simple task like detection can be reused by much complex tasks such as segmentation and tracking. The most significant benefit of the usage of transfer learning is obtaining a better model initialization, which is very important for training the model with limited samples.


Our proposed network taking use of data fusion transfer learning strategy. In specific, the designed model is pretrained on HSI datasets captured by different sensors with 3D pyramid pooling, and then fine-tuned on the target datasets to achieve a better performance.


\section{Methodology}

Among the deep learning models employed in the HSI literatures, 3D-CNNs are much suitable than 2D-CNNs for HSI classification due to fact that HSI with the 3D data format. In fact, different objects in HSI generally have different spectral structures. Convolving along the spectral dimension is very important. In addition, there are also some different objects which have similar spectral structures. For these objects, convolving along the spatial dimension to capture features is also beneficial. 

For 2D-CNNs based methods, on the one hand, without spectral dimension reduction, the number of parameters of 2D-CNNs will be extremely large due to that HSIs generally have hundreds of bands. On the other hand, if the dimension reduction is conducted, it may destroy the information of spectral structure which is important for discriminating different objects. 

Generally speaking, 3D-CNN based approaches have better performance than 2D-CNN based approaches \cite{chen2016deep,zhong2018spectral}. However, despite being much accurate, the existing 3D-CNN based approaches still have two shortcomings. 1) Comparing with 2D convolutions, 3D convolutions have more parameters and 3D CNN models are computation-intensive. 2) Being limited by the training samples in HSI datasets, 3D-CNN models employed in HSI classification almost consist of less than five convolution layers. Although a large number of experiments in computer vision have proved that deep depth of CNN is very important for improving the performance of tasks related to image processing.

In this section, we start with the description of the proposed AINet, and end with a introduction to the proposed data fusion transfer learning strategy.

\subsection{AINet for HSI Classification} 
\begin{figure}[t]
\setlength{\abovecaptionskip}{0.cm}
\setlength{\belowcaptionskip}{-0.cm}
\centering
	\includegraphics[height=4cm]{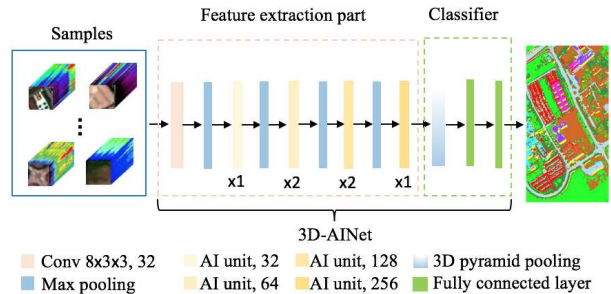}
	\caption{Framework of AINet. On the left, the $L\times S\times S$-sized samples from the neighborhood window centered around each target pixel are extracted first, and then the samples are fed to AINet to extract deep spectral-spatial features. Finally, the classification scores are calculated by the classifier.}
	\label{fig:AINet}
\end{figure}

\noindent
\textbf{Network Structure} 
Figure~\ref{fig:AINet} shows the overall framework of the proposed AINet for HSI classification. In order to fully utilize the spectral and spatial informations contained in HSI, we extract $L\times S\times S$-sized cubes from raw HSI as samples, where $L$ and $S$ respect the number of spectrum bands and the spatial size accordingly (Following \cite{chen2016deep}, we set $S$ to 27 in this paper). Then the samples are fed to AINet to extract deep spectral-spatial features and to calculate classification scores. Inspired by the design of ResNet \cite{he2016deep}, AINet employs a similar basic structure and introduces some key modifications for tailoring on HSI dataset. AINet starts with a 3D convolution layer, then stacks six AI units with increasing widths, and connects one 3D spatial pyramid pooling and one fully connected layer at the end. Specifically, the channels for the six AI units are 32, 64, 64, 128, 128 and 256 accordingly. In order to reduce the dimension of features, four Max pooling layers are added with kernel=[3, 3, 3], stride=[2, 2, 2] within the six AI units. 

\vspace{2mm}
\noindent
\textbf{3D Pyramid Pooling} 
Before the fully connected layer, there is a 3D pyramid pooling for mapping features with different sizes to vectors with fixed dimensions. Different HSI datasets are usually captured by different sensors and with various numbers of spectrum bands, for example, the Pavia University dataset has 103 bands and the Indian Pines dataset contains 200 bands. With 3D pyramid pooling layer, the same network can be applied on different HSI datasets without any modification. In this paper, 3D spatial pyramid pooling layer is composed of three level pooling ($1\times 1 \times 1$, $2\times 1\times 1$, $3\times 1\times 1$). As the last AI unit has 256 channels, the outputs of 3D spatial pyramid pooling layer are $256\times 6\times 1\times 1$-sized cubes. 

\vspace{2mm}
\noindent
\textbf{Training and Loss} 
We employ log\_softmax \cite{de2015exploration} as the activation function in the fully connected layer. During training, we take negative log likelihood as the loss function, and add ${L}_{2}$ regularization term with weight $1e-5$ to the loss function for alleviating over-fitting. And the optimizer is stochastic gradient descent (SGD) with momentum \cite{krizhevsky2012imagenet}. For all of the experiments, the same setting is adopted, where momentum, weight decay, batch size, epochs and learning rate are  0.9, 1e-5, 20, 60 and 0.01 respectively. During the last 12 epochs, the learning rate is decreased to 0.001.

\subsection{AI Unit}

\begin{figure}[t]
\setlength{\abovecaptionskip}{0.cm}
\setlength{\belowcaptionskip}{-0.cm}
	\centering
	\includegraphics[height=2.4cm]{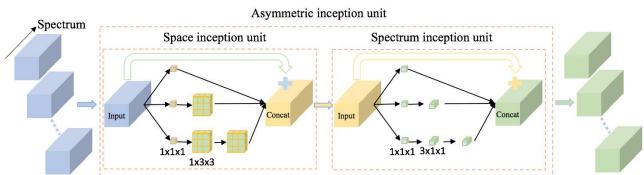}
	\caption{Illustration of a AI unit. In AI unit, 3D convolution layer is replaced with two asymmetric inception units, that are space inception unit and spectrum inception unit. In space inception unit, the input cube is fed to three different paths. With path one has a point wise convolution layer only, path two consists of one point wise convolution layer and one 2D convolution layer, and path three has one point wise convolution layer and two 2D convolution layers. The outputs of each path are concatenated in channel, and are added to the output of the shortcut connection. The structure of spectrum inception unit is similar with the space inception unit, except that $1\times 3\times 3$-sized convolution layers are replaced with $3\times 1\times 1$-sized convolution layers in spectrum inception unit.}
	\label{fig:AIUnit}
\end{figure}

Due to 3D convolution can learn spectral and spatial information from raw HSI dataset, 3D-CNN based methods can obtain the state-of-the-art performance for HSI classification. However, 3D convolutions are prone to over-fitting and computation-intensive compared with the 2D convolutions. In order to address these problem, we propose the asymmetric inception unit (AI unit), which is consist of the space inception unit and the spectrum inception unit. The structure of AI unit is illustrated in Figure~\ref{fig:AIUnit}. In the space inception unit, there are three space convolution paths. Path one has one point wise convolution layer only, path two consists of one point wise convolution layer and one 2D convolution layer with $1\times 3 \times 3$-sized kernels, and path three has one point wise convolution layers and two 2D convolution layers. The outputs of each path are concatenated in channel, and are added to the output of the shortcut connection. Inspired by the Inception networks \cite{szegedy2017inception}, we set the three paths with different widths. For each unit, we set the widths of three paths with a split ratio 1:2:1. In the last two paths, the width of point wise convolution layer is half of that of the other convolution layers. For instance, in the AI unit with 32 channels, the width of the first path is 8, and for the second path, the widths of the point wise convolution layer and $1\times 3\times 3$-sized convolution layer are 8 and 16 respectively , the widths of the three layers of the last path are 4, 8 and 8 accordingly. In the overall structure, the structure of spectrum inception unit is similar with the space inception unit, except that $1\times 3\times 3$-sized 2D convolution layers in space inception unit are replaced with $3\times 1\times 1$-sized 1D convolution layers.

\begin{figure}[t]
\setlength{\abovecaptionskip}{0.cm}
\setlength{\belowcaptionskip}{-0.cm}
	\centering
	\includegraphics[height=4.6cm]{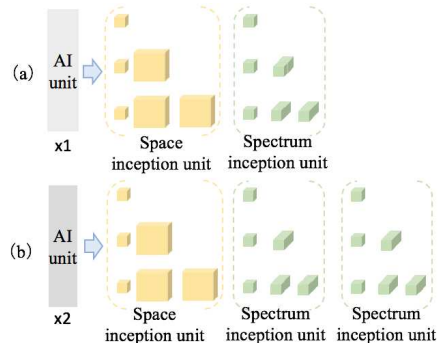}
	\caption{Illustration of stacking two AI units. (a) AI unit $\times$1; (b) AI unit $\times$2. Instead of stacking two AI units with the same type, we stack one space inception unit and two spectrum inception units to form AI unit $\times$2 as shown in (b).}
    \vspace{-0.1cm}
    \label{fig:AIunit_2}
\end{figure}

In HSI datasets, the resolution of spectrum is much higher than that of space and the information of spectrum is much richer. Therefore, during spectral-spatial features extraction, we are paying more attention on spectral feature extraction. In the proposed AINet, there are six AI units. The four units located in the middle can be divided into two groups and each group stacks two units with equal width. Here, instead of stack two same AI units in each group, we stack one space inception unit and two spectrum inception units. This is different from some popular networks, such as ResNet\cite{he2016deep} and MobileNet\cite{howard2017mobilenets}, which build the whole model by stacking same units. Figure~\ref{fig:AIunit_2} shows the difference between one AI unit and two AI units.

\subsection{Transfer Learning with Data Fusion}

\begin{figure}[h]
	\centering
    \includegraphics[height=8cm]{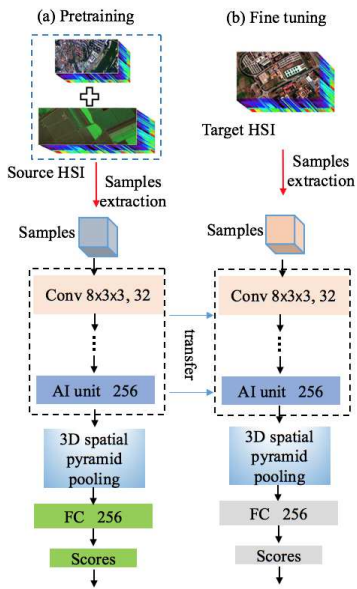}
	\caption{Data fusion transfer learning strategy. (a) Data fusion pretraining. During pertraining, the proposed network are trained on two different HSI datasets for improving the diversity of samples and obtaining a good initialized model. (b) Fine tuning. After acquiring the pretrained model, a new model is initialized with the parameters of pretrained model for the target HSI dataset.}
	\label{fig:transfer}
\end{figure}

In RGB image classification, pretraining large-scale network on ImageNet dataset which has over 14 millions of hand-annotated images and over 20 thousands of categories is common, and it is very useful for improving the performance and overcoming the problem of limited training samples. During transfer learning, the diversity of the dataset used for pretraining is the key factor. For example, pretraining the same model on the dataset which has million images with one thousand categories always achieving better performance than the one pretraining on the dataset which has 10 million images with ten categories. We suspect that the model pertraining with the more diverse samples may acquire a better generalization ability.

In HSI classification, the labeled samples are quite limited. However, all of the HSI datasets just contain a few categories. For further improving the performance of HSI classification, we propose a data fusion transfer learning strategy. As shown in Figure~\ref{fig:transfer}, the strategy is composed of pretraining and fine tuning. During pretraining, the proposed network are trained on two source HSI datasets. Here, Pavia Center dataset and Salinas dataset are used as source HSI datasets for pretraining. Among the several public HSI datasets, those two datasets have the largest number of labeled samples. To be more specific, the model is initialized with Gaussian distribution on one source HSI dataset and pretrained for $N$ epochs, and then the feature extraction part is fixed and the classifier is reinitialized with Gaussian distribution. Later on, the feature extraction part and classifier on the other source HSI dataset are pretrained for $\frac{N}{2}$ epochs with a different learning rate. In this paper, $N$ is set to 10 and the learning rate used for the feature extraction part is tenth of that used for the second pretraining HSI dataset.

After the model is pretrained on the two source HSI datasets, we transfer the whole mode except the classifier to the fine-tuning model built for the target HSI dataset as initialization, and then fine tuning the transfer part and new classifier with the same learning rate as the one used for training the second source HSI dataset.

\subsection{Experiments}

\subsection{Datasets and Experiments Setting}

\begin{figure}
	\centering
    \includegraphics[height=5.6cm]{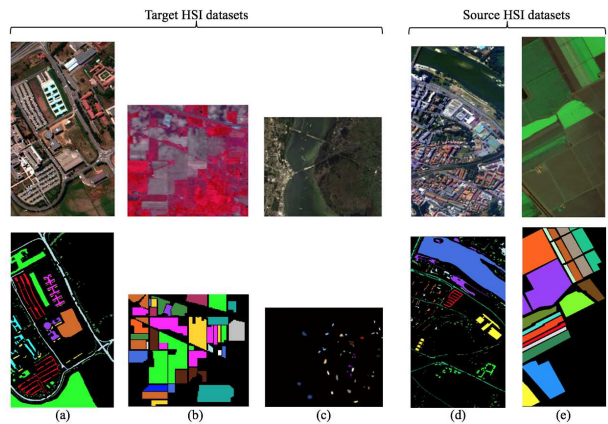}
	\caption{False-color composites (first row) and ground truths (second row) of experimental HSI datasets. Each color represents one kind of object. (a) Pavia University; (b) Indian Pines; (c) Kennedy Space Center; (d) Pavia Center; (e) Salinas.}
	\label{fig:datasets}
\end{figure}

In this paper, we compare the proposed AINet with other four CNN based approaches for HSI classification on three public HSI datasets, including Pavia University, Indian Pines and KSC. For the experiments with transfer learning, Pavia Center dataset and Salinas dataset are employed as the source datasets. The false-color composite and ground truth of each dataset are shown in Figure~\ref{fig:datasets}. A brief introduction of each dataset is given in the following part and more information can be found on the website\footnote{\url{http://www.ehu.eus/ccwintco/index.php}}.

Pavia University and Pavia Center datasets are captured by Reflective Optics System Imaging Spectrometer (ROSIS) sensor in 2001. After several noisiest bands being removed, Pavia University has 103 bands and Pavia Center has 102 bands. Both datasets are divided into 9 classes. Indian Pines and Salinas datasets are acquired by the Airborne Visible/Infrared Imaging Spectrometer (AVIRIS) sensor in 1992. After correction, each dataset has 200 bands and contains 16 classes. KSC is acquired by the AVIRIS sensor in 1996, and after removing water absorption and low SNR bands, 176 bands were used for analysis. For classification purposes, 13 classes are defined.

For the three target HSI datasets, samples are divided into training samples and testing samples. For comparison purposes, we follow \cite{chen2016deep} to set the samples distribution for Indian Pines and KSC datasets. And for the Pavia University dataset, we extract 200 samples in a random way from each class as training samples. 

In the experiments using transfer learning, we randomly extract 200 samples from each class of Pavia Center dataset, 100 samples from each category of Salinas dataset as test samples, and taking the rest as training samples. 

\subsection{Performance Comparison of different network structures}

\setlength{\tabcolsep}{4pt}
\begin{table}
\setlength{\abovecaptionskip}{0.cm}
\setlength{\belowcaptionskip}{-0.cm}
\renewcommand{\arraystretch}{1}
\begin{center}
\caption{
Classification results for the Pavia University dataset.
}
\label{table:headings_1}
\begin{tabular}{cccccc}
\hline\noalign{\smallskip}
models &1D-CNN & 2D-CNN & 3D-CNN & SSRN & AINet \\
\hline
\# train & 3930 & 3930 & 3930 & \bf{1800} & \bf{1800} \\
\# param. & \bf{2898} & 0.183M & 5.849M & 0.453M & 0.487M\\
depth & 4 & 4 & 4 & 12 & \bf{32} \\
\hline
OA & 92.28 & 94.04 & \bf{99.54} &  98.98 & 99.42\\
AA & 92.55 & 97.52 & \bf{99.66} & 99.07 &  99.51\\
$K$ & 90.37 & 92.43 & \bf{99.41} & 98.64 & 99.22 \\
\hline
\end{tabular}
\end{center}
\end{table}
\setlength{\tabcolsep}{1.4pt}

\setlength{\tabcolsep}{4pt}
\begin{table}[h]
\setlength{\abovecaptionskip}{0.cm}
\setlength{\belowcaptionskip}{-0.cm}
\renewcommand{\arraystretch}{1}
\begin{center}
\caption{
Classification results for the Indian Pines dataset.
}
\label{table:headings_2}
\begin{tabular}{cccccc}
\hline\noalign{\smallskip}
models & 1D-CNN & 2D-CNN & 3D-CNN & SSRN & AINet \\
\hline
\# train & 1765 & 1765 & 1765 & 1765 & 1765 \\
\# param. & \bf{25920} & 0.183M & 44.893M & 0.453M & 0.487M\\
depth &6 & 4 & 4 & 12 & \bf{32} \\
\hline
OA & 87.81 & 89.99 & 97.56 &  98.40 & \bf{99.14}\\
AA & 93.12 & 97.19 & 99.23 & 98.52 & \bf{99.47}\\
$K$ & 85.30 & 87.95 & 97.02 & 98.14 & \bf{99.00}\\
\hline
\end{tabular}
\end{center}
\end{table}
\setlength{\tabcolsep}{1.4pt}

\setlength{\tabcolsep}{4pt}
\begin{table}
\setlength{\abovecaptionskip}{0.cm}
\setlength{\belowcaptionskip}{-0.cm}
\renewcommand{\arraystretch}{1}
\begin{center}
\caption{
Classification results for the KSC dataset.
}
\label{table:headings_3}
\begin{tabular}{cccccc}
\hline\noalign{\smallskip}
models & 1D-CNN & 2D-CNN & 3D-CNN & SSRN & AINet \\
\hline
\# train & 459 & 459 & 459 & 459 & 459 \\
\# param. & \bf{14904} & 0.183M & 5.849M & 0.453M & 0.487M\\
depth & 5 & 4 & 4 & 12 & \bf{32} \\
\hline
OA & 89.23 & 94.11 & 96.31 &  98.65 &  \bf{99.01}\\
AA & 83.32 & 91.98 & 94.68 & 97.78 &  \bf{98.65}\\
$K$ & 86.91 & 93.44 & 95.90 & 98.54 &  \bf{98.90}\\
\hline
\end{tabular}
\end{center}
\end{table}
\setlength{\tabcolsep}{1.4pt}

In this section, we compare the proposed AINet with other four CNN based HSI classification methods, that are 1D-CNN, 2D-CNN, 3D-CNN \cite{chen2016deep}, SSRN \cite{zhong2018spectral}, on three aforementioned HSI datasets. The experiments with same settings are running for 5 times to acquire the average performance. The experimental results are listed in Tables 1$\sim$3, where the number of training samples, the number of parameters used in the convolution layers, the depth of CNN models, overall accuracy (OA), average accuracy (AA) and kappa coefficient ($K$) are reported. OA is the ratio between the number of correctly classified samples in test set and the total number of test set. AA is the mean of the OA of all the categories. $K$ is a coefficient which measures inter-rater agreement for qualitative items \cite{thompson1988reappraisal}.    

From tables 1$\sim$3, we can see that, the proposed AINet achieves best classification performance on all of the datasets. For instance, in Indian Pines dataset, OA of AINet is 99.14, which is 9.15\% better than that of 2D-CNN, 1.58\% better than that of 3D-CNN and 0.74 better than that of SSRN. Through the experiments, it is easy to find that all of the 3D-CNN based HSI classification methods obtained better performance than 2D-CNN. From 3D-CNN, SSRN to AINet, the depth of the models are increasing and the classification accuracy goes up. In specific, the depths of the three models are 4, 12, 32 accordingly. Although AINet is much deeper than SSRN in depth, the parameters of AINet is slightly more than that of SSRN and much less than that of 3D-CNN. 

\subsection{Results of Transfer Learning}

\begin{figure}[h]
	\centering
    \includegraphics[height=2.8cm]{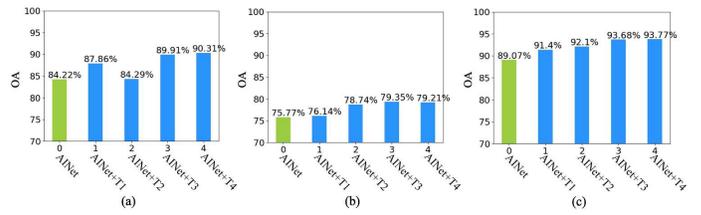}
	\caption{Transfer learning experiments with 15 training samples per class. (a) Pavia University; (b) Indian Pines; (c) Kennedy Space Center.}
	\label{fig:15samples}
\end{figure}

\begin{figure}[h]
	\centering
    \includegraphics[height=2.8cm]{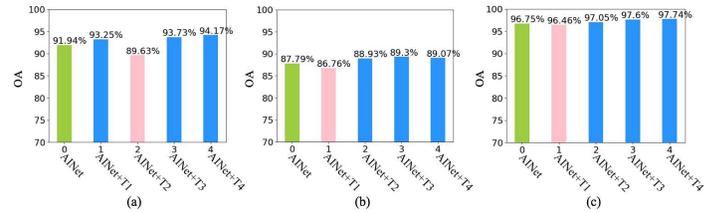}
	\caption{Transfer learning experiments with 30 training samples per class. (a) Pavia University; (b) Indian Pines; (c) Kennedy Space Center.}
	\label{fig:30samples}
\end{figure}

In this part, we combine the proposed AINet with data fusion transfer learning to further improve the classification performance. For each dataset, we choose $\{15, 30\}$ samples from each class as training samples to test the affect brought by the number of the training samples. The experiment results using transfer learning are shown in Figure~\ref{fig:15samples} and Figure~\ref{fig:30samples}, where AINet represents our model trained only on the target dataset, AINet+T1 represents our model pretrained on Pavia Center, which captured from the same sensor as Pavia University dataset, and fine-tuning on the target dataset. AINet+T2 represents our model pretrained on Salinas dataset, which captured from the same sensor as the Indian Pines dataset, and fine-tuning on the target dataset. AINet+T3 and AINet+T4 are our models pretrained on both Pavia Center dataset and Salina datasets, with different orders accordingly, and fine-tuning on the target dataset.

From the overall trends of Figure~\ref{fig:15samples} and Figure~\ref{fig:30samples}, we can draw two conclusions. Firstly, the performance of our AINet is benefit from the transfer learning, especially when the available training samples are relatively small, the performance gains are huge. Secondly, pretrained on two different datasets achieve much gains than the ones just pretrained on a single dataset. Which we infer the benefits mainly come from the various classes and categories of the pretrain datasets. 

From Figure~\ref{fig:15samples}, there are two points that can be concluded: Firstly, from the performance of AINet+T, AINet+T1, AINet+T2 in (a),(b),(c), we can conclude that AINet pretrained on other HSI dataset, will lead performance increasing for the target HSI dataset, no matter if the source and target dataset are captured by the same sensor or not. Secondly, from the performance of AINet+T, AINet+T1, AINet+T2 in (a),(b), we can conclude that when pretraining on other HSI dataset, the performance gains brought by the same sensor is much larger than the one brought by different sensors for small training samples.

From Figure~\ref{fig:30samples}, when the training samples become larger, the performance of HSI classification is still benefit from transfer learning. But notice that the pink bars in Figure~\ref{fig:30samples}, which have a decreasing in performance, the reason behind that is still unclear and need the  further investigation. One possible reason we suspect is that on the pretrained dataset, a different learning rate is used comparing with the one used for fine-tuning on the target dataset, and which may not lead to the exactly same converge direction for the training process between the fine-tuning dataset and target dataset.

\section{Conclusions}
\label{sec:conclusions}

This paper proposes a 3D asymmetric inception network for hyperspectral image classification named AINet.
Firstly, AINet using a 3D CNN light-weight while still very deep, which can exert the potential of the deep learning for extracting the representative features, and meanwhile alleviate the problem brought by the limited annotation datasets. Secondly, considering the property of hyperspectral image, spectral signatures are emphasized than spatial contexts. Moreover, data fusion transfer learning strategy is utilized for a better model initialization and saving the training time. In the future, there are two topics we are keen to pursue, to investigate the reduction of the training time brought by transfer learning is the first one, and another one is taking use of some policies to overcome the data imbalance in HSI classification. 

\bibliographystyle{named}
\bibliography{reference}

\end{document}